\documentclass[10pt,journal,compsoc]{IEEEtran}
\ifCLASSOPTIONcompsoc
  \usepackage[nocompress]{cite}
\else
  \usepackage{cite}
\fi
\ifCLASSINFOpdf
\else
\fi

\usepackage{subfigure}
\usepackage{graphicx}
\usepackage{algorithmic}
\usepackage[algoruled,vlined]{algorithm2e}
\usepackage{amsmath}
\usepackage{amssymb}
\usepackage{url}

\begin{document}
%
\title{Wasserstein CNN: Learning Invariant Features for NIR-VIS Face Recognition}
%
%
%
%

\author{Ran~He,~\IEEEmembership{Senior Member,~IEEE,}
        Xiang~Wu,
        Zhenan~Sun$^*$,~\IEEEmembership{Member,~IEEE,}
        and~Tieniu~Tan,~\IEEEmembership{Fellow,~IEEE}
\IEEEcompsocitemizethanks{\IEEEcompsocthanksitem $^*$Zhenan~Sun is the corresponding author. R. He, X. Wu, Z. Sun and T. Tan are with National Laboratory of Pattern Recognition, CASIA, Center for Research on Intelligent Perception and Computing, CASIA, Center for Excellence in Brain Science and Intelligence Technology, CAS and University of Chinese Academy of Sciences, Beijing, China,100190.\protect\\
E-mail: \{rhe, znsun, tnt\}@nlpr.ia.ac.cn, xiang.wu@cripac.ia.ac.cn} }

%
%

\markboth{Journal of \LaTeX\ Class Files,~Vol.~14, No.~8, August~2017}%
{Shell \MakeLowercase{\textit{et al.}}: Bare Demo of IEEEtran.cls for Computer Society Journals}
%



\IEEEtitleabstractindextext{%
\begin{abstract}

Heterogeneous face recognition (HFR) aims to match facial images acquired from different sensing modalities with mission-critical applications in forensics, security and commercial sectors. However, HFR is a much more challenging problem than traditional face recognition because of large intra-class variations of heterogeneous face images and limited training samples of cross-modality face image pairs. This paper proposes a novel approach namely Wasserstein CNN (convolutional neural networks, or WCNN for short) to learn invariant features between near-infrared and visual face images (i.e. NIR-VIS face recognition). The low-level layers of WCNN are trained with widely available face images in visual spectrum. The high-level layer is divided into three parts, i.e., NIR layer, VIS layer and NIR-VIS shared layer. The first two layers aim to learn modality-specific features and NIR-VIS shared layer is designed to learn modality-invariant feature subspace. Wasserstein distance is introduced into NIR-VIS shared layer to measure the dissimilarity between heterogeneous feature distributions. So W-CNN learning aims to achieve the minimization of Wasserstein distance between NIR distribution and VIS distribution for invariant deep feature representation of heterogeneous face images. To avoid the overfitting problem on small-scale heterogeneous face data, a correlation prior is introduced on the fully-connected layers of WCNN network to reduce parameter space. This prior is implemented by a low-rank constraint in an end-to-end network. The joint formulation leads to an alternating minimization for deep feature representation at training stage and an efficient computation for heterogeneous data at testing stage. Extensive experiments on three challenging NIR-VIS face recognition databases demonstrate the significant superiority of Wasserstein CNN over state-of-the-art methods.
\end{abstract}

}

\maketitle

\IEEEdisplaynontitleabstractindextext

%
\IEEEpeerreviewmaketitle

\IEEEraisesectionheading{\section{Introduction}\label{sec:introduction}}

\IEEEPARstart{U}{biquitous} face sensors not only facilitate the wide application of face recognition but also generate various heterogeneous sets of facial images~\cite{RHe:2017_1}\cite{SOuyang:2016}. Matching faces across different sensing modalities raises the problem of heterogeneous face recognition (HFR) or cross-modality face
recognition. Due to significant difference in sensing processes, heterogeneous images of the same subject have a large appearance variation, which has distinguished HFR from regular visual (VIS) face recognition~\cite{DGong:2017}. During the last decade, HFR has become increasingly important in many practical security applications and drawn much attention in the computer vision community. Impressive progress has been made in research areas such as near infrared (NIR) vs. VIS~\cite{SLi:2013}, sketch vs. VIS~\cite{XTang:2002}, 2D vs. 3D~\cite{ZLei:2008r}, different resolutions~\cite{Biswas:2012} and poses~\cite{RHuang:2017}, etc. 

Since NIR imaging technique provides an efficient and straightforward solution to improve face recognition performance in extreme lighting conditions, it has been considered as one of the most prominent alternative sensing modalities in HFR~\cite{Reale:2016}.
Moreover, NIR imaging has been proved to be less sensitive to visible light illumination variations \cite{JZhu:2014}, and thus is applicable to face recognition at a distance or even at night-time. It has been widely used in face identification or authorization applications, such as security surveillance and E-passport. However, most face galleries only consist of VIS images due to the mass deployment of VIS sensors, while the probe images often come in NIR modalities. Therefore, the demand for robust matching between NIR and VIS face images, also known as the NIR-VIS heterogeneous face recognition problem, has greatly raised and drawn much attention.

Much research effort has been made to improve the NIR-VIS HFR performance~\cite{SOuyang:2016}\cite{XXLiu1:2016}. Traditional NIR-VIS methods generally involve image synthesis, subspace learning and invariant feature extraction~\cite{YJin:2015}\cite{SOuyang:2016}. These methods are often based on several processing steps to achieve satisfying accuracy. Recently, inspired by the successful application of convolutional neural networks (CNN) in VIS face recognition \cite{YSun:2014}\cite{Taigman:2014}\cite{Schroff:2015}, several deep models~\cite{XXLiu:2016}\cite{Saxena:2016}\cite{Reale:2016} attempt to transfer the knowledge learned on a large scale VIS face database to NIR modality. These methods firstly train a basic CNN network on the public CASIA NIR-VIS 2.0 database~\cite{SLi:2013} and then make the basic network adaptable to both NIR and VIS modalities. Experimental results suggest that deep models have a potential to outperform the traditional NIR-VIS methods.

However, NIR-VIS HFR still remains a challenging problem for deep models and is largely unsolved mainly due to the following two reasons: {\bf 1) The gap between sensing patterns of VIS and NIR modalities.} Since NIR and VIS images are captured from different sensing modalities, they have large differences in feature representations. Lacking representative spectral information of NIR images, the deep models trained on VIS data fail to provide satisfying results~\cite{XXLiu:2016}\cite{Saxena:2016}\cite{Reale:2016}.
The debate on the optimal measurement of the difference and approach to close the gap between VIS and NIR modalities remains active, and thus it is still challenging in exploring modality-invariant representations of both NIR and VIS face images via large-scale VIS face data.
{\bf 2) The over-fitting on small-scale training set.} With the thriving development of Internet, large collection of VIS face images can be gathered more efficiently.
However, VIS face images paired with NIR layout can hardly be available online, making paired VIS and NIR images expensive to obtain at large scale.
Most existing HFR databases are of small-scale (fewer than 10,000 samples) while having large feature dimensions (at least 100$\times$100 pixels).
Consequently, deep models will likely to over-fit to the training set during feature learning~\cite{XXLiu:2016}\cite{Reale:2016}. Exploring the optimal method to fit deep models to small-scale NIR-VIS datasets remains a central problem.

In this paper, the two aforementioned problems are tackled by a novel Wasserstein CNN (WCNN) architecture. WCNN employs one single network structure to map both NIR and VIS images to a compact Euclidean feature space so that the NIR and VIS images in the embedding space directly correspond to face similarity. WCNN is composed of three key components in an end-to-end fashion. First of all, inspired by the observation and results that the appearance of a face is composed of identity information and variation information (e.g., lightings, poses, and expressions)~\cite{DChen:2012}\cite{SLi:2013}\cite{DYi:2015}, we divide the high-level layer of WCNN into two orthogonal subspaces that contain modality-invariant identity information and modality-variant spectrum information, respectively. Secondly, we focus on the way to evaluate how close the NIR distribution and the VIS distribution are. Wasserstein distance is imposed on the identity subspace to measure the difference between NIR and VIS feature distributions, which reduces the gap between the two modalities. The learned identity subspace is expected to contain the identity invariant information of the two modalities. We further assume that the features of the same subject in the identity subspace follow a Gaussian distribution so that the Wasserstein distance can be efficiently optimized. Lastly, considering that the fully connected layers of WCNN have a large number of parameters and are prone to over-fit on small-scale dataset, we impose a correlation prior on the fully connected layers, which is implemented by a non-convex low-rank constraint. The advantage of this prior is particularly significant when a training set is small.

Our convolutional network is first trained on large-scale VIS data. Its convolutional layers and fully connected layer are implemented by the simplest case of maxout operator~\cite{Goodfellow:2013}. This network makes our learned representation to be robust to intra-class variations of individuals. Then, the low-level layers of this network are fine-tuned to be adaptable to NIR data. Our joint formulation leads to an alternating minimization approach for deep representation at the training time and an efficient computation for heterogeneous data at the testing time. The effectiveness of our WCNN method is extensively evaluated using the most challenging CASIA NIR-VIS 2.0 Database~\cite{SLi:2013}, Oulu-CASIA NIR-VIS Database~\cite{JChen:2009} and BUAA NIR-VIS Database~\cite{DHuang:2012}. Our results demonstrate that the proposed WCNN method clearly outperforms the related state-of-the-art NIR-VIS methods, and significantly improve state-of-the-art rank-1 accuracy and verification rate (VR) at a low false acceptance rate (FAR).

The main contributions of our work are summarized as follows,
\begin{itemize}
\item An effective end-to-end network architecture is developed to learn modality invariant features. This architecture could naturally combine invariant feature extraction and subspace learning into a unified network. Two orthogonal subspaces are embedded to model identity and spectrum information respectively, resulting in one single network to extract both NIR and VIS features.
\item A novel Wasserstein distance is introduced to measure the distribution difference between NIR and VIS modalities. Compared to previous sample-level measures \cite{XXLiu:2016}\cite{Reale:2016}, Wasserstein distance could effectively reduce the gap between the two modalities and results in better feature representation.
\item A correlation prior is imposed on the fully connected layers of deep models to alleviate the over-fitting problem on small scale datasets. This prior makes the proposed WCNN work well on small-scale NIR-VIS dataset and significantly improves verification rate on a low verification rate.
\item Experimental results on the challenging CASIA NIR-VIS 2.0 face database show that WCNN advances the best verification rate (@FAR=0.1\%) from 91.0\% to 98.4\%. Compared with state-of-the-art results~\cite{XLiu:2016}, it further reduces the error rate (1-VR) by 82\% only with a compact 128-D feature representation.
\end{itemize}

The rest of this paper is organized as follows. We briefly review some related work on NIR-VIS heterogeneous face recognition in Section~\ref{sec:rw}. In Section~\ref{sec:IDR}, we present the details of our Wasserstein CNN approach for NIR-VIS face recognition. Section~\ref{sec:exp} provides experimental results, prior to summary in Section~\ref{sec:con}.



\section{Related work \label{sec:rw}}
The problem of heterogeneous identity matching across different sensing modalities has received increasing attention in biometrics community. Almost all types of biometrics (e.g., face and iris~\cite{LXiao:2013}) have encountered this problem. NIR-VIS HFR has been one of the most extensively researched subject in heterogeneous biometrics. We briefly describe some recent works on this related subject and generally categorize these works into four classes~\cite{JZhu:2014}\cite{YJin:2015}\cite{DGong:2017}: image synthesis, subspace learning, feature representation and deep learning. 

{\bf Image synthesis} methods aim to synthesize face images from one modality (or domain) into another so that heterogeneous images can be compared in the same distance space. These methods try to handle the difference of sensing modalities at image preprocessing stage. Image synthesis was firstly used in face photo-sketch synthesis and recognition~\cite{XTang:2003}. \cite{RWang:2009} applied face analogy to transform a face image from one modality to another. \cite{XWang:2009} resorted to multiscale Markov random fields to synthesize pseudo-sketch to face photo. Then, \cite{XGao:2008} further used hidden Markov model to learn the nonlinear relationship between face photos and sketches. \cite{ZLei:2008r} reconstructed a 3D face model from a single 2D face image using canonical correlation analysis (CCA). \cite{SWang:2012}, \cite{DHuang:2013} and \cite{FXu:2015} used coupled or joint dictionary learning to reconstruct face images and then performed face recognition. Recently, a cross-spectral hallucination and low-rank embedding was proposed in \cite{Lezama:2017} to synthesize a VIS image from a NIR image in a patch way. Although better rank-1 accuracy was claimed in \cite{Lezama:2017}, \cite{Lezama:2017} does not follow the standard 10-fold testing protocol~\cite{SLi:2013}. Since image synthesis is an ill-posed problem and a photo-realistic synthesis image is usually difficult to generate, this kind of approaches can only reduce the modality difference to some extent~\cite{DGong:2017}.

{\bf Feature representation} methods try to explore modality-invariant features that are robust to various sensing conditions. The current methods are almost based on hand-crafted local features, such as local binary patterns (LBP), histograms of oriented gradients (HOG), Difference-of-Gaussian (DoG) and SIFT \cite{SLiao:2009}\cite{Klare:2010}\cite{Goswami:2011}. In addition, \cite{LHuang:2012} applied sparse representation to learn modality-invariant features. \cite{Klare:2011} further applied the densely sampled SIFT and multi-block LBP features to represent heterogeneous face images. \cite{JZhu:2014} combined Log-DoG filtering, local encoding and uniform feature normalization together to find better feature representation. Based on bag of visual words, \cite{MShao:2016} proposed a hierarchical hyperlingual-words to capture high-level semantics across different modalities. \cite{DGong:2017} converted face images pixel by pixel into encoded face images with a trained common encoding model, and then applied a discriminant method to match heterogeneous face images. Feature extraction methods reduce the modality difference when converting heterogeneous images to features, and are often applied along with subspace learning methods.

{\bf Subspace learning} methods learn mappings to project homogenous data into a common space in which inter-modality difference is minimized as much as possible. CCA and partial least squares (PLS) are two representative methods. \cite{DLin:2006} proposed a common discriminant feature extraction approach to incorporate both discriminative and local information. \cite{ZLei:2012} developed a coupled discriminant analysis based on the locality information in kernel space. \cite{XHuang:2013} proposed a regularized discriminative spectral regression method to map heterogeneous data into a common spectral space. Recently, \cite{KWang:2013} took feature selection into consideration during common subspace learning. \cite{Klare:2013} proposed prototype random subspace method with kernel similarities for HFR. State-of-the-art NIR-VIS results are often obtained by removing some principal subspace components \cite{DYi:2015}. Multi-view discriminant analysis~\cite{MKan:2016} and mutual component analysis~\cite{ZFLi:2016} were further developed to reduce the modality difference.

{\bf Deep learning} methods mainly resort to CNN to extract deep feature representation of heterogeneous images. These methods are often pre-trained on a large-scale VIS dataset, and then are fine-tuned on NIR face images to learn a modality invariant representation. \cite{Saxena:2016} used a pre-trained VIS CNN along with different metric learning strategies to improve HFR performance. \cite{XXLiu:2016} employed two types of NIR-VIS triplet loss to reduce intra-class variations and to augment the number of training sample pairs. \cite{Reale:2016} trained two networks (named VisNet and NIRNet) with small convolutional filters, and coupled the two networks' output features by creating a Siamese network with contrastive loss. By performing CNN, these methods achieved a verification rate of 91.03\% at FAR of 0.1\% and rank-1 accuracy of 95.74\% on the challenging CASIA NIR-VIS 2.0 database~\cite{XXLiu:2016}. However, compared to VIS recognition, the performance of NIR-VIS HFR is still far from satisfying.
For example, rank-1 accuracy on the CASIA NIR-VIS 2.0 face database is significantly lower than that on the Labeled Faces in the Wild (LFW) VIS database \cite{GHuang:2007} (Rank-1 accuracy has been more than 99\%). The high performance of VIS recognition benefits from deep learning techniques and large amounts of VIS face images. However, due to the gap and over-fitting problem, NIR-VIS HFR is still challenging for deep learning methods.

The invariant deep representation method was first proposed in our early work~\cite{RHe:2017}. Apart from providing more in-depth analysis and more extensive experiments, the major difference between this paper and \cite{RHe:2017} is the introduction of the new Wasserstein distance and correlation constraint. Our experiments suggest that the new Wasserstein distance could better measure the feature distribution difference between NIR and VIS face data, leading to further improvement of recognition performance (especially in a lower false acceptance rate). In addition, the correlation constraint on the fully connected layers of WCNN could make learned features more adaptable to small-scale NIR training database, which also improves the performance. Compared with our early work~\cite{RHe:2017}, our new WCNN method reduces the error rate by 62\% at FAR=0.1\%.

\begin{figure*}[t]
\centering
\includegraphics[width=132mm]{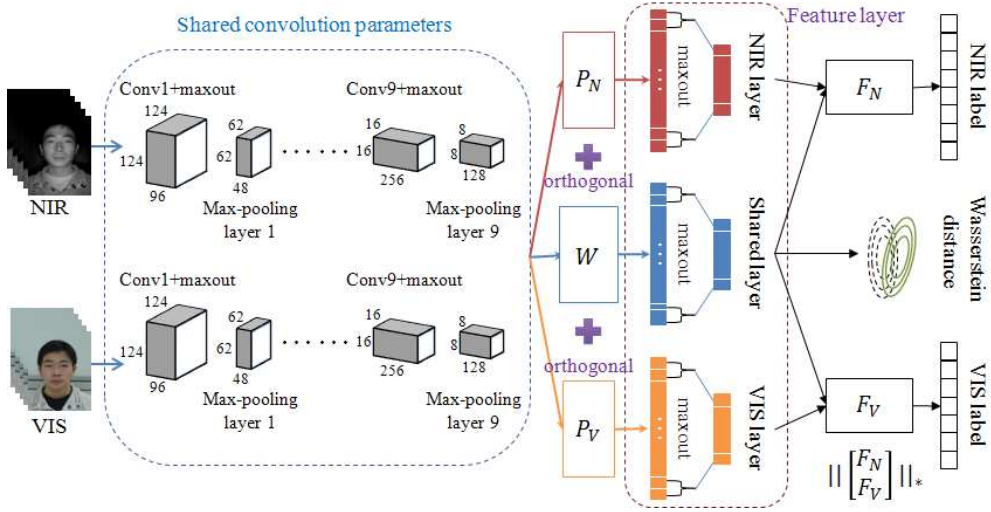}
\caption{An illustration of our proposed Wasserstein CNN architecture. The Wasserstein distance is used to measure the difference between NIR and VIS distributions in the modality invariant subspace (spanned by matrix $W$). At the testing time, both NIR and VIS features are exacted from the shared layer of one single neural network and compared in cosine distance. \label{fig:illu}}
\end{figure*}

\section{The proposed Wasserstein CNN \label{sec:IDR}}
Benefiting from the development of convolutional neural network (CNN), VIS face recognition has made great progress in recent years~\cite{YSun:2014}\cite{Taigman:2014}\cite{Schroff:2015}. This section introduces a new CNN architecture to learn modality invariant deep features for NIR-VIS HFR, named Wasserstein CNN, which consists of three key components as shown in Fig.~\ref{fig:illu}. The first component aims to seek a low-dimensional subspace that contains modality-invariant features. The second one explores the Wasserstein distance to measure the difference between NIR and VIS distributions. The last one imposes correlation prior on the fully connected layers to alleviate over-fitting on small-scale NIR dataset.

\subsection{Problem Formulation}
Let $I_V$ and $I_N$ be the VIS and NIR images respectively. The CNN feature extraction process is denoted as $X_i = Conv(I_i, \Theta_i)$ ($i \in \{N,V\}$), where $Conv()$ is the feature extraction function defined by the ConvNet, $X_i$ is the extracted feature vector, and $\Theta_i$ denotes ConvNet parameters for modality $I$ to be learned. In heterogeneous recognition, one basic assumption is the fact that there is some common concepts between heterogeneous samples. Hence, we assume that NIR and VIS face images share some common low-level features. That is, $\Theta_N = \Theta_V = \Theta$ and $X_i = Conv(I_i, \Theta)$. As shown in Fig.~\ref{fig:illu}, the output of the last max-pooling layer represents $X_i\in \mathrm{R}^p$, corresponding to the NIR and VIS channel, respectively. These two channels share the same parameter $\Theta$.

{\bf Modality Invariant Subspace:} Previous NIR-VIS matching methods often use a trick to alleviate the problem of appreance variation by removing some principal subspaces that are assumed to contain light spectrum information \cite{SLi:2013}\cite{DYi:2015}. Observation and results also demonstrate that the appearance of a face is composed of identity information and variation information (e.g., lightings, poses, and expressions) \cite{DChen:2012} and removing spectrum information is helpful for NIR-VIS performance~\cite{DYi:2015}. Inspired by these results, we introduce three mapping matrices (i.e., $W, P_i\in \mathrm{R}^{d\times p}$) in CNN to model identity invariant information and variant spectrum information. Therefore, the deep feature representation can be defined as
\begin{equation}
f_i=\left[
\begin{array}{l}
f_\mathrm{shared}\\
f_\mathrm{unique}
\end{array}
\right]
=\left[
\begin{array}{l}
WX_i\\
P_iX_i
\end{array}
\right] \; (i \in \{N,V\}),
\end{equation}
where $WX_i$ and $P_iX_i$ denote the shared feature and the unique feature respectively. Considering the subspace decomposition properties of the matrices $W$ and $P_i$, we further impose an orthogonal constraint to make them to be unrelated to each other, i.e.,
\begin{equation}
P_i^TW=0 \; (i \in \{N,V\}).
\end{equation}
This orthogonal constraint could also reduce parameter space and alleviate over-fitting. Different from previous methods \cite{DYi:2015}\cite{ZFLi:2016}\cite{MShao:2016}\cite{DGong:2017} that treat feature representation and subspace learning as two independent steps, our architecture is able to naturally combine these two steps in an end-to-end network.

{\bf The Wasserstein Distance:} The gap of sensing mechanism between NIR and VIS images is a major difficulty in HFR. Previous methods often resort to sample-level constraints to reduce this gap. The triplet loss and contrastive loss are imposed on NIR-VIS sample pairs in \cite{XXLiu1:2016} and \cite{Reale:2016} respectively. These methods only consider the relationship between NIR-VIS samples rather than NIR-VIS distributions. Recently, Wasserstein distance proves to play a prominent role of measuring the model distribution and the real distribution in generative adversarial networks (GAN)~\cite{Arjovsky:2017}\cite{Berthelot:2017}. Inspired by Wasserstein GAN~\cite{Arjovsky:2017} and BEGAN~\cite{Berthelot:2017}, we make use of Wasserstein distance to measure how close NIR data distribution and VIS data distribution are. Considering that NIR-VIS data are from different subjects and there are large extra-class variations, we impose Wasserstein distance on the distributions of one subject. We further assume the data distributions of one subject follow a Gaussian distribution after non-linear feature mapping. The Gaussian distribution assumption in Wasserstein distance have been shown to be effective in image generation problem~\cite{Berthelot:2017} and sequence matching problem~\cite{BSu:2017}. Experimental results show that this assumption also provides meaningful learning results for HFR.

Given the two Gaussian distributions $X=\mathcal{N}(m_N, C_N)$ and $Y=\mathcal{N}(m_N, C_N)$ corresponding to one subject, where the means $m_N, m_V\in \mathbb{R}^p$ and the covariances $C_N, C_V\in \mathbb{R}^{p\times p}$, the 2-Wasserstein distance between $X$ and $Y$ of one subject could be defined as~\cite{Berthelot:2017}:
\begin{equation} \label{eq:w2}
W_2(X,Y)^2=\|m_N - m_V\|_2^2 + \text{trace}(C_N+C_V - 2(C_V^{\frac{1}{2}}C_NC_V^{\frac{1}{2}})^{\frac{1}{2}}).
\end{equation}
As in~\cite{Berthelot:2017}, we simplify (\ref{eq:w2}) to:
\begin{equation}
\begin{split}
W_2(X,Y)^2 &=\frac{1}{2}\left[\|m_N - m_V\|_2^2 + (c_N+c_V-2\sqrt{c_Nc_V})\right] \\
 &=\frac{1}{2}\left[\|m_N - m_V\|_2^2 + \|\sigma_N-\sigma_V\|_2^2\right],
\end{split}
\end{equation}
where the $\sigma_N$ and $\sigma_V$ are the standard deviations of $X$ and $Y$, taking the following forms:
\begin{equation}
\begin{split}
\sigma_1 = \sqrt{\frac{1}{n}\sum_{i=0}^n(x_i-m_N)^2}=\sqrt{\frac{1}{n}\sum_{i=0}^nx_i^2-m_N^2}, \\
\sigma_2 = \sqrt{\frac{1}{n}\sum_{i=0}^n(y_i-m_V)^2}=\sqrt{\frac{1}{n}\sum_{i=0}^ny_i^2-m_V^2}.
\end{split}
\end{equation}
Their gradients can be computed as
\begin{equation}
\frac{\partial W_2}{\partial x_i}  = \frac{1}{n}(m_N-m_V) + (\sigma_1-\sigma_2)\frac{\partial{(\sigma_N-\sigma_V)}}{\partial x_i},
\end{equation}
where
\begin{equation}
\frac{\partial{(\sigma_N-\sigma_V)}}{\partial x_i} =  \frac{2}{n}\frac{(x_i-m_N)}{\sqrt{\sigma_V^2+\epsilon}},
\end{equation}
and $\epsilon$ is a constant. Therefore, the final gradient of $X$ can be denoted as
\begin{equation}
\frac{\partial W_2}{\partial x_i} = \frac{1}{n}\left[(m_N-m_V)+ 2(\sigma_N-\sigma_V)\frac{(x_i-m_N)}{\sqrt{\sigma_N^2+\epsilon}}\right].
\end{equation}
Analogously, the gradient of $Y$ can be written as
\begin{equation}
\frac{\partial W_2}{\partial y_i} = -\frac{1}{n}\left[(m_N-m_V)+ 2(\sigma_N-\sigma_V)\frac{(y_i-m_V)}{\sqrt{\sigma_V^2+\epsilon}}\right].
\end{equation}


{\bf Correlation Prior:} One challenge of applying CNN to HFR is the over-fitting problem of CNN on a small-scale training set. In CNN, fully connected layers often take up the majority of the parameters. Since there are both NIR and VIS labels in HFR, the number of class labels in HFR is twice larger than that in VIS face recognition. A large number of class labels also result in fully connected layers of large size. Hence, when the training set is of small-scale, fully connected layers can not be well adjusted and are easy to be over-fitting. The fully connected layer of WCNN is composed of two matrices $F_N$ and $F_V$ corresponding to NIR and VIS modalities respectively. We expected that $\mathrm{M}=\begin{bmatrix}F_N\\F_V\end{bmatrix}$ are highly correlated so that $M^TM$ is a block-diagonal matrix~\footnote{Block-diagonal prior was used in subspace segmentation to make clustering results more accurately~\cite{JFeng:2014}. It requires an affinity matrix to be block-diagonal to characterize sample clusters}. A correlated $M$ will reduce the estimated parameter space and naturally alleviate the over-fitting problem. We make use of the matrix nuclear norm on $\mathrm{M}$, i.e.,
\begin{equation}
\|\mathrm{M}\|_*=\text{tr}(\sqrt{\mathrm{M}^T\mathrm{M}}).
\end{equation}
The matrix nuclear norm requires that $M$ has a low-rank structure and its elements are linearly correlated. Then $M^TM$ tends to be a block-diagonal matrix. Given the SVD decomposition of $\mathrm{M}=U\Sigma V^T$, we can obtain:
\begin{equation}\label{nuclear_norm}
\mathcal{R} =\|\mathrm{M}\|_*
            =\text{tr}(\sqrt{V\Sigma U^TU\Sigma V^T})= \text{tr}(\sqrt{\Sigma^2}).
\end{equation}
Since the elements of $\Sigma$ are non-negative, the gradient of the nuclear norm can be written as:
\begin{equation}\label{gradient_F_1}
\frac{\partial\mathcal{R}}{\partial \mathrm{M}}=\frac{\partial{\text{tr}(\Sigma)}}{\partial \mathrm{M}}=UV^T.
\end{equation}
Therefore, we can use $UV^T$ as the subgradient of nuclear norm. Note that since the fully connected matrices $F_V$ and $F_N$ are not used in the testing time, the correlation prior only intends to alleviate over-fitting rather than compress a network.


\subsection{Optimization Method}
The commonly used softmax loss is used to train the whole network, taking the following form,
\begin{equation}
\begin{array}{c}
\begin{split}
\mathcal{L}_{cls}& =\sum_{i \in \{N,V\}}{\text{softmax}(F_i, c, \Theta, W, P_i)}\\
                               & =-\sum_{i \in \{N,V\}}(\sum_{j=1}^N \mathbf{1} \{y_{ij}=c\}\text{log}{\hat{p}_{ij}})\\
\end{split}\\
\\
\mathrm{s.t.} \qquad P_i^TW=0 \; (i \in \{N,V\})
\end{array}
\end{equation}
where $c$ is the class label for each sample and $\hat{p}_{ij}$ is the predicted probability. Moreover, we denote $\mathbf{1}\{\cdot\}$ as the indicator function so that $\mathbf{1}\{\text{a true statement}\}=1$ and $\mathbf{1}\{\text{a false statement}\}=0$.

According to the theory of lagrange multipliers, (\ref{eq:obj}) can be reformulated as an unconstrained problem, 
\begin{equation}
\begin{split}
\mathcal{L}_{cls}= &\sum_{i\in\{N,V\}}{\text{softmax}(F_i, c, \Theta, W, P_i)}\\
                                 & + \sum_{i\in\{N,V\}}\lambda_i\|P_i^TW\|_F^2,
\end{split}
\label{eq:obj_un}
\end{equation}
where $\lambda_i$ are the lagrange multipliers and $\|\cdot\|_F^2$ denotes the Frobenius norm.

To decrease the discrepancy between different modalities, we apply Wasserstein distance to measure the two distributions of NIR and VIS images from one subject.
\begin{equation}
\mathcal{L}_{dist} = \frac{1}{2}\left[\|m_N - m_V\|_2^2 + \|\sigma_N-\sigma_V\|_2^2\right].
\end{equation} 
Specially, under the WCNN training scheme, we employ mini-batch stochastic gradient descent to optimize the objective function, so the statistics of each mini-batch are used to represent the means and standard deviations instead.

To alleviate over-fitting, we also introduce Eq.(\ref{nuclear_norm}). Then the final objective function takes the following form,
%
\begin{equation}\label{eq:obj}
\mathcal{L} = \beta_1\mathcal{L}_{cls} + \beta_2\mathcal{L}_{dist} + \beta_3R + \sum_{i\in\{N,V\}}\lambda_i\|P_i^TW\|_F^2,
\end{equation}
where $\beta_1, \beta_2$ and $\beta_3$ are the trade-off coefficients for each part. If gradient descent method is used to minimize Eq.(\ref{eq:obj}), we should update the parameters $W, P_i, F_i$ and $\Theta$. For the convolutional parameters $\Theta$, we follow the back-propagation method to update it. The gradients of $W$, $P_i$ and $F_i$ can be expressed as
\begin{equation}
\frac{\partial \mathcal{L}}{\partial W}= \frac{\partial\mathcal{L}_{cls}}{\partial W} + \frac{\partial\mathcal{L}_{dist}}{\partial W}
\end{equation}
\begin{equation}
\frac{\partial \mathcal{L}}{\partial P_i}= \frac{\partial\mathcal{L}_{cls}}{\partial P_i} + \frac{\partial\mathcal{L}_{dist}}{\partial P_i}
\end{equation}
\begin{equation}
\frac{\partial \mathcal{L}}{\partial F_i}= \frac{\partial\mathcal{L}_{cls}}{\partial F_i} + \frac{\partial\mathcal{R}}{\partial F_i}
\end{equation}

\begin{algorithm}[tb]
\caption{Training the Wasserstein CNN network.}
\label{algorithm1}
\begin{algorithmic}[1]
\REQUIRE
Training set $X_i$, learning rate $\gamma$ and lagrange multipliers $\lambda_i$.
\ENSURE
The CNN parameters $\Theta$ and the mapping matrix $W$.
\STATE Initialize parameters $\Theta$ by pre-trained model and the mapping matrices $W, P_i, F_i$ by Eq.(\ref{eq_init});
\FOR {$t=1,\dots, T$}
\STATE CNN optimization:
\STATE \quad Update $\Theta, W, P_i, F_i$ via back-propagation method;
\STATE Fix $\Theta$:
\STATE \quad Update $W$ according to Eq.(\ref{gradient_W});
\STATE \quad Update $P_i$ according to Eq.(\ref{gradient_V});
\STATE \quad Update $F_i$ according to Eq.(\ref{gradient_F_1});
\ENDFOR;
\STATE \textbf{Return} $\Theta$ and $W$;
\end{algorithmic}
\end{algorithm}

Note that the updating gradients for $W$, $P_i$ and $F_i$ contain two parts. The first one is used for conventional back-propagation in CNN. The second part of $W, P_i$ for subspace learning can be re-organized in
\begin{equation}
\frac{\partial \mathcal{L}}{\partial W}= \sum_{i\in\{N,V\}}\lambda_iP_iP_i^TW
\label{gradient_W}
\end{equation}
\begin{equation}
\begin{split}
\frac{\partial \mathcal{L}}{\partial P_i}= \lambda_iWW^TP_i
\end{split}
\label{gradient_V}
\end{equation}
For the low-rank correlation constraint, we can update $M=[F_N, F_V]^T$ by Eq.(\ref{gradient_F_1}). Then we update these parameters with a learning rate $\gamma$ via
\begin{align}
\Theta^{(t+1)} &= \Theta^{(t)} -\gamma \frac{\partial \mathcal{L}}{\partial \Theta^{(t)}}\\
W^{(t+1)} &= W^{(t)} -\gamma \frac{\partial \mathcal{L}}{\partial W^{(t)}}\\
P_i^{(t+1)} &= P_i^{(t)} -\gamma \frac{\partial \mathcal{L}}{\partial P_i^{(t)}} \\
F_i^{(t+1)} &= F_i^{(t)} -\gamma \frac{\partial \mathcal{L}}{\partial F_i^{(t)}}
\end{align}

Since Eq.(\ref{eq:obj}) contains several variables and is non-convex, we develop an alternating minimization method to minimize Eq.(\ref{eq:obj}) in an end-to-end CNN optimization scheme. First, we update the parameters by conventional back-propagation to optimize CNN. Then, we fix the CNN parameters and update matrices $W, P_i, F_i$ by their own gradients. The optimization detail is summarized in Algorithm~\ref{algorithm1}. As in \cite{Xavier:2010}, the parameters $\Theta$ of CNN is initialized by the pre-trained model and the mapping matrices $W, P_i, F_i$ is initialized by
\begin{equation}
W, P_i, F_i \sim U\left[-\frac{1}{\sqrt{m}},\frac{1}{\sqrt{m}}\right]
\label{eq_init}
\end{equation}
where $U\left[-a,a\right]$ is the uniform distribution in the interval $(-a,a)$ and $m$ is the dimension of original features.

\subsection{Network Structure}
The basic VIS network architecture (the convolution parameters sharing part in Fig.~\ref{fig:illu}) and initial values of $\Theta$ are trained on a large-scale VIS dataset \cite{YGuo:2016}. We employ the light CNN network \cite{XWu:2015} as the basic network\footnote{\url{https://github.com/AlfredXiangWu/face_verification_experiment}}. The network includes nine convolution layers with four max-pooling layers, followed by the fully connected layer. Softmax is used as the loss function. The training VIS face images are normalized and cropped to $144\times144$ according to five facial points. To enrich the input data, we randomly cropped the input images into $128\times128$. The MS-Celeb-1M dataset~\cite{YGuo:2016}, which contains totally 8.5M images for about 100K identities, is employed to train the basic network. Dropout ratio is set to $0.7$ for fully connected layer and the learning rate is set to $1e^{-3}$ initially and reduced to $1e^{-5}$ for $4,000,000$ iterations. The trained single model for the basic network obtained $98.90\%$ on the LFW dataset.

Based on the basic VIS network, we develop a modality invariant convolution neural network for NIR-VIS face recognition. The low-level convolution layers are initialized by the pre-trained basic network. We implement two CNN channels with shared parameters to input NIR and VIS images respectively. Then we define the feature layer (as in Fig.~\ref{fig:illu}) that aims to project the low-level features into two orthogonal feature subspaces. In this way, we can leverage the correlated properties of NIR and VIS identities and enforce the domain-specific properties of both modalities. When the summation of Wasserstein distance over all subjects reaches zero, invariant deep features are learned. Finally, the softmax loss functions are separately used for NIR and VIS representation as the supervisory signals. Note that since there is a maxout operator in the feature layer, the final feature dimension is $d/2$ when $W \in \mathrm{R}^{d\times m}$. As in VIS training, all NIR and VIS images are cropped and resized to $144 \times 144$ pixels and a randomly selected $128\times 128$ regions are fed into WCNN for NIR-VIS training. The learning rate of the Wasserstein CNN is set to $1e^{-4}$ initially and reduced to $1e^{-6}$ gradually for around $100,000$ iterations. The trade-off parameters $\beta_1, \beta_2$ and $\beta_3$ can be set to $1, 1$ and $0.001$, respectively.

\begin{figure*}[t]
    \subfigure[The CASIA NIR-VIS 2.0 database]{\includegraphics[height=24mm]{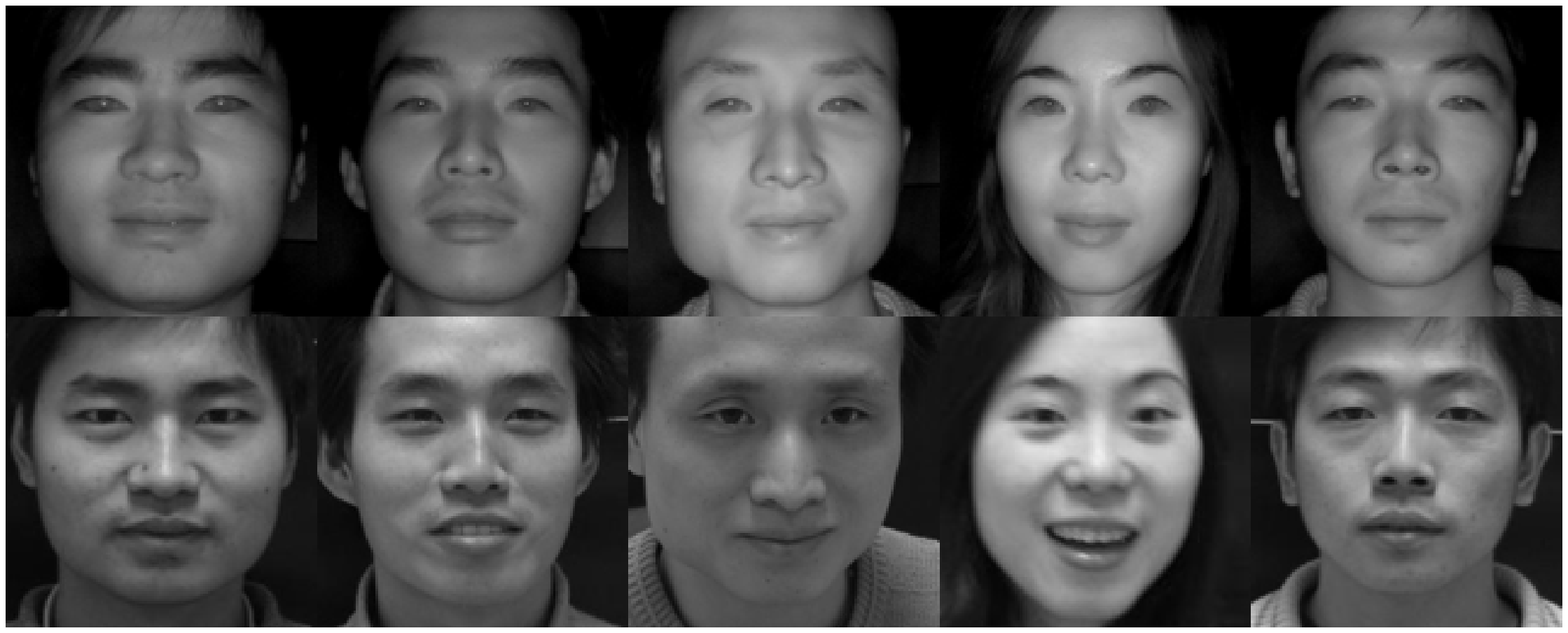}}
    \subfigure[The Oulu-CASIA NIR-VIS database]{\includegraphics[height=24mm]{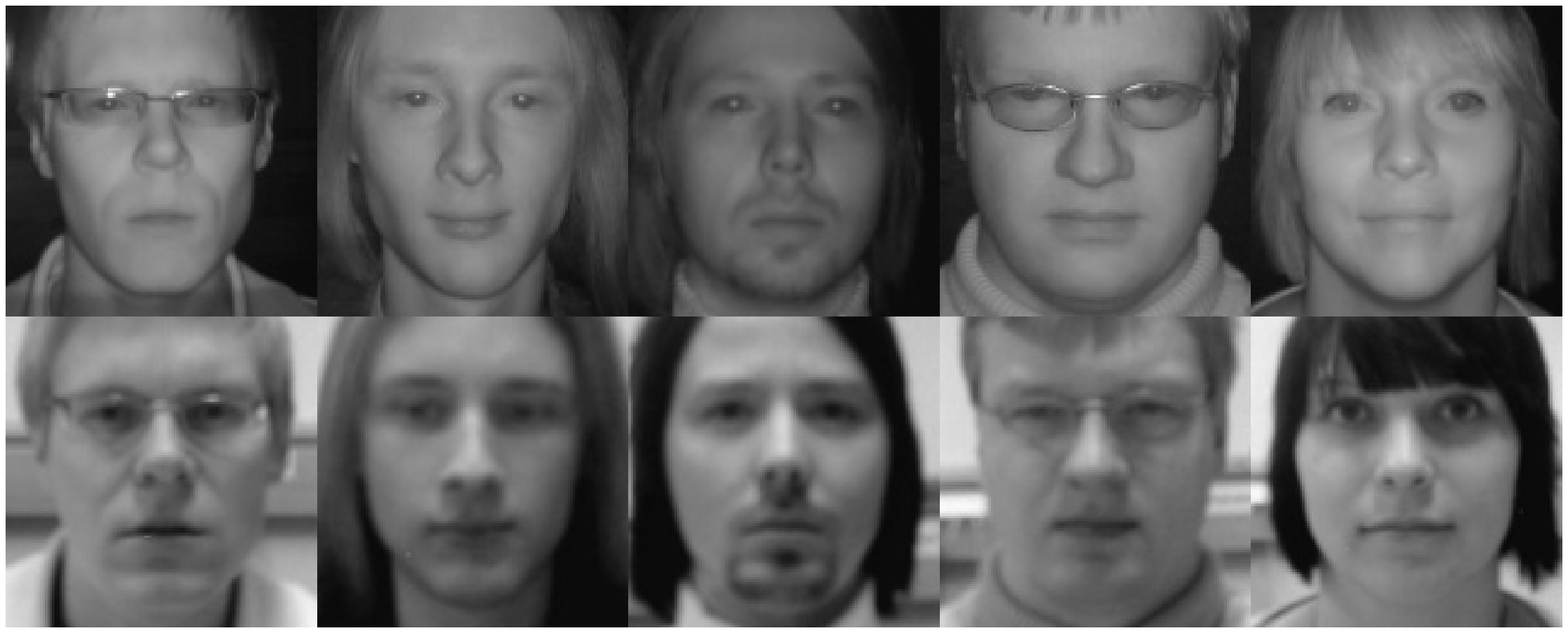}}
   \subfigure[The BUAA-VisNir database]{\includegraphics[height=24mm]{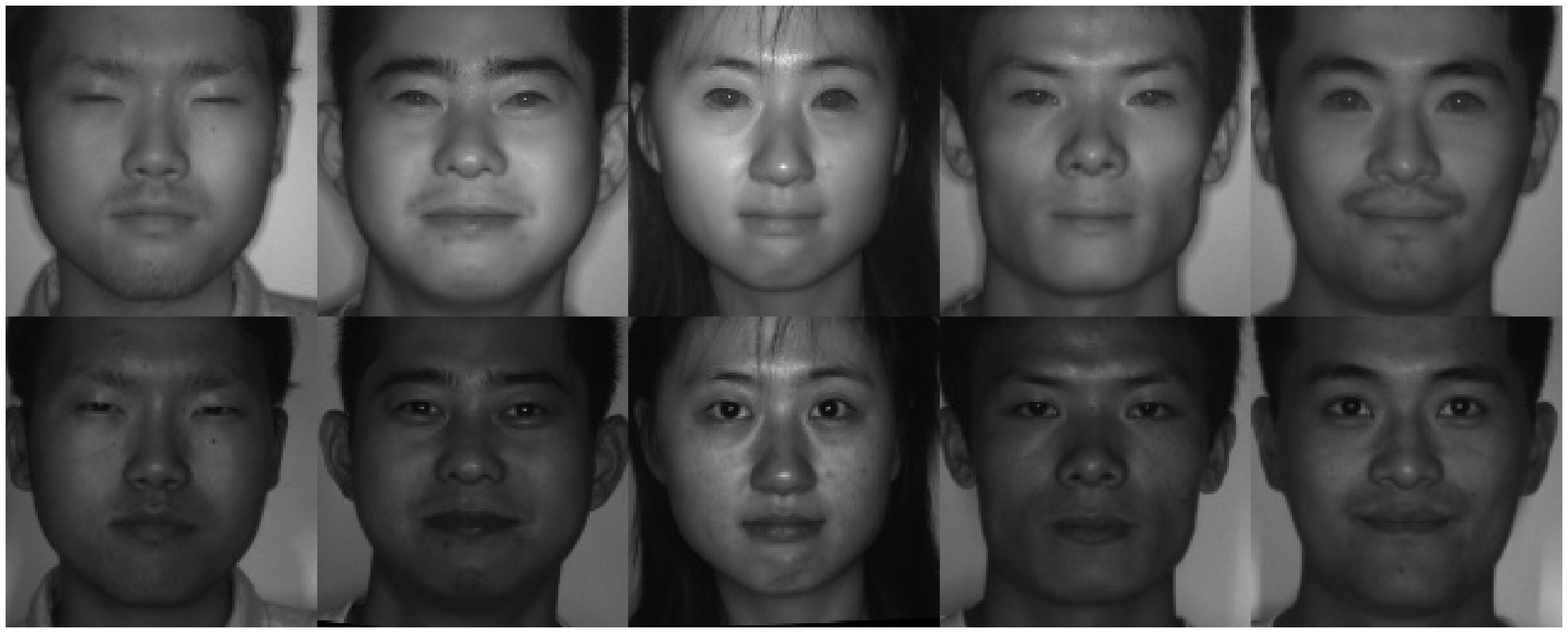}}
   \caption{Cropped VIS and NIR facial images in the three databases. The first row contains the NIR images from the probe set and the second row contains the VIS images from the gallery set. \label{fig:face}}
\end{figure*}
\section{Experiments and Results \label{sec:exp}}
In this section, we systemically evaluate the proposed WCNN approach against traditional methods and deep learning methods on three recently published NIR-VIS face databases: CASIA NIR-VIS 2.0 database, Oulu-CASIA NIR-VIS database and BUAA-VisNir database. Fig.~\ref{fig:face} shows the samples of cropped VIS and NIR facial images in the three databases.

\subsection{Datasets and Protocols}
{\bf The CASIA NIR-VIS 2.0 Face Database}~\cite{SLi:2013} is widely used in NIR-VIS heterogeneous face evaluations because it is the largest public and most challenging NIR-VIS database. Its challenge is due to large variations of the same identity, including lighting, expression, pose, and distance. Wearing glasses or not is also considered to generate variations. The database is composed of 725 subjects, each with 1-22 VIS and 5-50 NIR images. Each image is randomly gathered so that there are not one-to-one correlations between NIR and VIS images. The database contains two views of evaluation protocols. View 1 is used for super-parameters adjustment, and View 2 is used for training and testing.

For a fair comparison with other results, we follow the standard protocol in View 2. There are 10-fold experiments in View 2. Each fold contains a collection of training and testing lists. Nearly equal numbers of identities are included in the training and testing sets, and are kept disjoint from each other. Training on each fold is many-to-many (i.e., images from NIR and VIS are randomly combined). For each training fold, there are approximately 2,500 VIS images and 6,100 NIR images from around 360 subjects. These subjects are mutually exclusive from the 358 subjects in the testing set. That is, the subjects in the training set and testing set are entirely different. The training set in each fold is used for IDR training. For each testing fold, the gallery set always contains a total of 358 subjects, and each subject only has one VIS image. The probe set has over 6,000 NIR images from the same 358 subjects. All the probe set is to be matched against the gallery set, resulting in a similarity matrix of size $358$ by around $6,000$.

{\bf The Oulu-CASIA NIR-VIS database}~\cite{JChen:2009} is composed of 80 subjects with six expression variations (anger, disgust, fear, happiness, sadness, and surprise). 50 subjects are from Oulu University and the remaining 30 subjects are from CASIA. Since the facial images of this database are captured under different environments from two institutes, their illumination conditions are slightly different~\cite{MShao:2016}. Following the protocols in \cite{MShao:2016}, we select a subset of this database for our experiments, including 10 subjects from Oulu University and 30 subjects from CASIA. Eight face images from each expression are randomly selected from both NIR and VIS. As a result, there are totally 96 (48 NIR images and 48 VIS images) images for each subject. 20 subjects are used as training and the remaining 20 subjects are used as testing. All VIS images of the 20 subjects in testing are as the gallery and all their corresponding NIR images are as the probe.

{\bf The BUAA-VisNir face database}~\cite{DHuang:2012} is often used for domain adaptation evaluation across imaging sensors. It has 150 subjects with 9 VIS images and 9 NIR images captured simultaneously. The nine images of each subject correspond to nine distinct poses or expressions: neutral-frontal, left-rotation, right-rotation, tilt-up, tilt-down, happiness, anger, sorrow and surprise. The training set and testing set are composed of 900 images of 50 subjects and 1800 images from the remaining 100 subjects respectively. As in~\cite{JChen:2009}, to avoid that the probe and gallery images are in the same pose and expression, only one VIS image of each subject is selected in the gallery set during testing. Hence, the gallery set and the probe set have 100 VIS images and 900 NIR images respectively. This testing protocol is challenging due to large pose and illumination variations in the probe set.

\subsection{Results on the CASIA NIR-VIS 2.0 Database}


\begin{table}[!t]
\renewcommand\arraystretch{1.3}
\caption{Rank-1 accuracy and verification rate on the CASIA 2.0 NIR-VIS face database. \label{tab:casia}}
\centering
\begin{tabular}{|l|c|c|c|c|}
\hline
Methods & Rank-1&FAR=1\%&FAR=0.1\%& Dim \\ \hline\hline
KCSR \cite{ZLei:2009}  & 33.8 & 28.5 & 7.6 & - \\ \hline
KPS \cite{Klare:2013}  & 28.2 & 17.4 & 3.7 & - \\ \hline
KDSR \cite{XHuang:2013}& 37.5 & 33.0 & 9.3 & -\\ \hline
PCA+Sym+HCA \cite{SLi:2013} & 23.7 & - &19.3 & - \\ \hline
LCFS \cite{KWang:2013}\cite{YJin:2015} & 35.4& 35.7& 16.7 & - \\ \hline
H2(LBP3) \cite{MShao:2016}& 43.8 & 36.5 & 10.1 &-\\ \hline
C-DFD \cite{ZLei:2014}\cite{YJin:2015} & 65.8 & 61.9& 46.2 & -\\ \hline
DSIFT\cite{Dhamecha:2014} & 73.3 &-& - &-\\ \hline
CDFL \cite{YJin:2015} & 71.5 & 67.7 &55.1 & $1000$\\ \hline
Gabor+RBM \cite{DYi:2015} & 86.2 &-& 81.3 &-\\ \hline
Recon.+UDP \cite{FXu:2015} & 78.5 &-& 85.8 & $ 1024$ \\ \hline
CEFD\cite{DGong:2017} & 85.6 & - & - &- \\ \hline\hline
VGG \cite{Parkhi:2015} &62.1 & 70.9 &39.7& 4096\\ \hline
SeetaFace \cite{XLiu:2016} &68.0 & 85.2 &58.8& 2048\\ \hline
TRIVET\cite{XXLiu:2016} & 95.7 & 98.1 &91.0 & 512 \\ \hline
HFR-CNNs \cite{Saxena:2016} &85.9 & - & 78.0 & - \\ \hline
IDNet \cite{Reale:2016} &87.1 & - & 74.5 & 320\\ \hline
IDR \cite{RHe:2017} & 97.3 & 98.9 & 95.7 & 128 \\\hline
WCNN  & 98.4 & 99.4& 97.6 & 128 \\\hline
WCNN + low-rank & \textbf{98.7} & \textbf{99.5} & \textbf{98.4} & 128 \\\hline
\end{tabular}
\end{table}

To verify the performance of IDR, we compare our method with state-of-the-art NIR-VIS recognition methods, including traditional methods and deep learning methods. Since most of methods follow the standard protocol to evaluate their performance on the CASIA NIR-VIS 2.0 database, we directly report their results from the published papers.

\begin{figure*}[t]
\center
    \subfigure[The CASIA NIR-VIS 2.0 database]{\includegraphics[width=60mm]{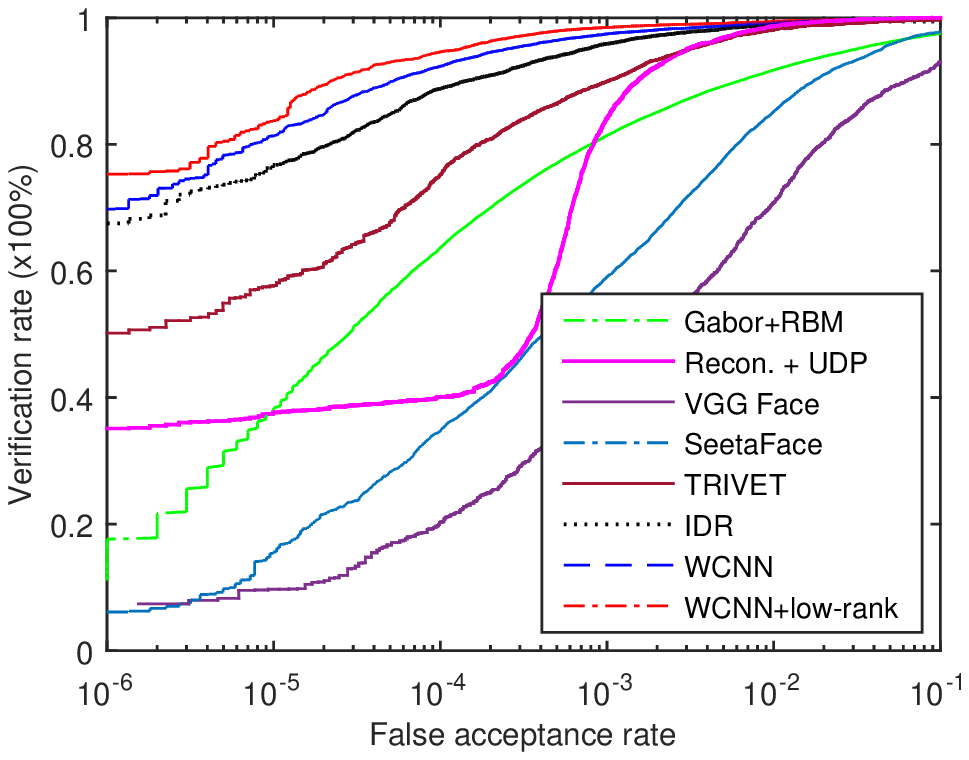}}
    \subfigure[The Oulu-CASIA NIR-VIS database]{\includegraphics[width=60mm]{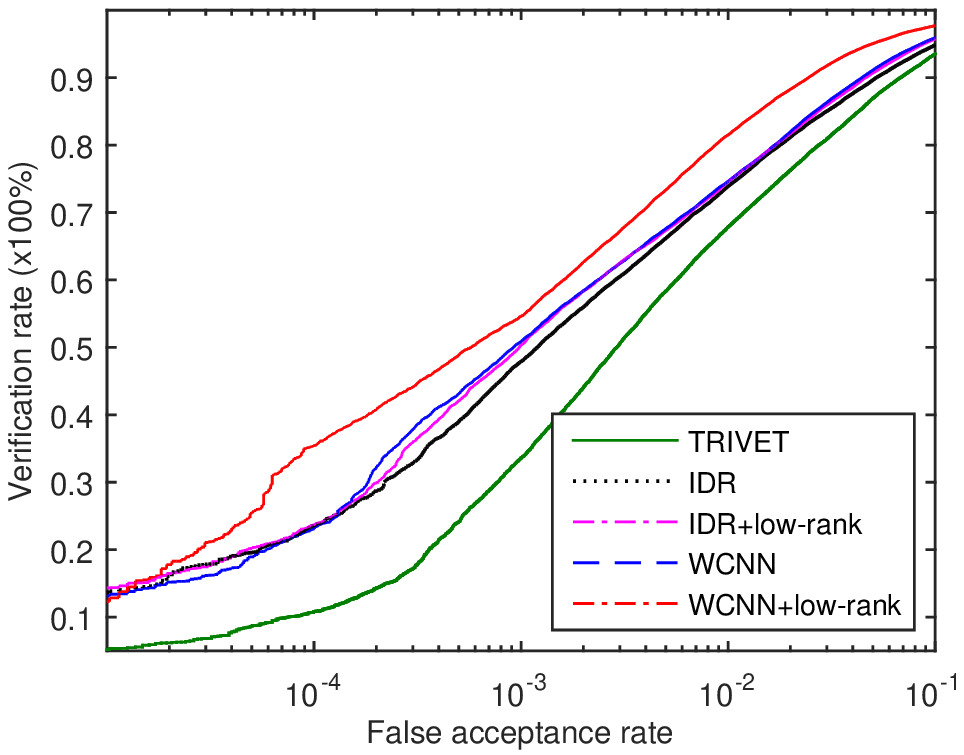}}
   \subfigure[The BUAA-VisNir database]{\includegraphics[width=60mm]{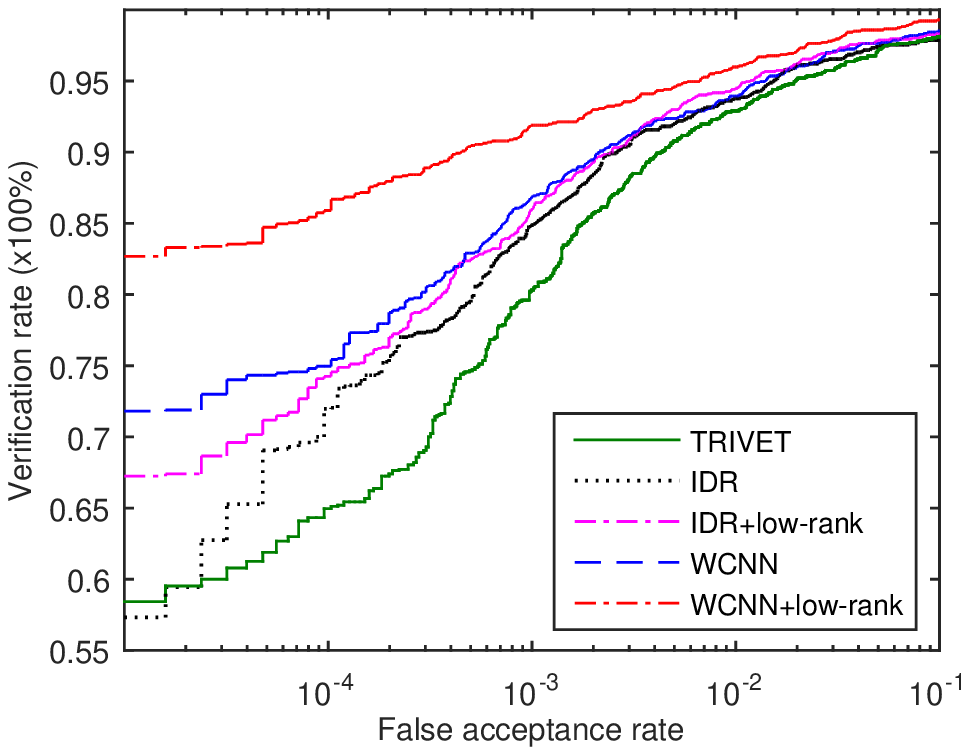}}
   \caption{ROC curves of different methods on the three NIR-VIS datasets. \label{fig:roc}}
\end{figure*}

The traditional methods include kernel coupled spectral regression (KCSR) \cite{ZLei:2009}, kernel prototype similarities (KPS) \cite{Klare:2013}, kernel discriminative spectral regression (KDSR) \cite{XHuang:2013}, PCA+Sym+HCA~\cite{SLi:2013}, learning coupled feature spaces (LCFS) \cite{KWang:2013}, coupled discriminant face descriptor (C-DFD) \cite{ZLei:2014}, DSIFT+PCA+LDA \cite{Dhamecha:2014}, coupled discriminant feature learning (CDFL) \cite{YJin:2015}, Gabor+RBM+Remove 11PCs \cite{DYi:2015}, reconstruction+UDP \cite{FXu:2015}, H2(LBP3) \cite{MShao:2016}, common encoding feature discriminant (CEFD)~\cite{DGong:2017}. The results of LCFS, C-DFD and CDFL are from \cite{YJin:2015}, and those of the remaining compared methods are from their published papers. For deep learning methods, we compare the recently proposed TRIVET\cite{XXLiu:2016}, HFR-CNNs \cite{Saxena:2016} and IDNet \cite{Reale:2016}. In addition, the results of two VIS CNN methods are also discussed, including VGG \cite{Parkhi:2015} and SeetaFace \cite{XLiu:2016}.

Table~\ref{tab:casia} shows the rank-1 accuracy and verification rates of different NIR-VIS methods. Fig.~\ref{fig:roc} (a) further plots the receiver operating characteristic (ROC) curves of the proposed method and its three top competitors. For a better illustration, we do not report some ROC curves of other methods if these curves are low. We have the following observations:

Due to the sensing gap, three VIS deep models can not work well for NIR-VIS HFR. The rank-1 accuracy and VR@FAR=0.1\% of VGG and SeetaFace are lower than those of state-of-the-art traditional methods, and significantly worse than those of the deep learning methods trained on NIR-VIS dataset. Compared with  VGG and SeetaFace, CEFD and Gabor+RBM can also obtain higher rank-1 accuracy. These results suggest that although large-scale VIS dataset is helpful for VIS face recognition, it has limited benefit for HFR if there is only a small-scale NIR dataset. Hence it is necessary to design suitable deep structures for NIR and VIS modalities. Then deep learning based methods (TRIVET, HFR-CNNs and IDNet) begin to outperform the traditional methods.

Compared to the traditional methods (CEFD, Gabor+RBM and reconstruction+UDP), the improvements of the recently proposed deep learning methods (TRIVET, HFR-CNNs and IDNet) are limited. Particularly, high rank-1 accuracy can not ensure a high verification rate or a better ROC curve.
Experimental results clearly show that our WCNN methods yield superior overall performance compared to other NIR-VIS methods. It is worth pointing out that one of the main strengths of WCNN is that it yields consistent improvement over rank-1 accuracy and verification rates. The advantage of WCNN is particularly apparent when FAR is low. Moreover, since we make use of orthogonal subspace to separate spectral information and identity information, the feature dimension of our method is smaller than that of other methods. All of these results suggest that deep learning is effective for the NIR-VIS recognition problem, and a compact and modality invariant feature representation can be learned from a single CNN.

Compared with our early version IDR~\cite{RHe:2017}, the WCNN+low-rank method further improves rank-1 accuracy from 97.3\% to 98.7\% and VR@FAR=0.1\% from 95.7\% to 98.4\%. It further reduces error rate (1-VR) by 62\% at FAR=0.1\%. Although rank-1 accuracy and VR@FAR=0.1\% of WCNN are high, the low-rank constraint could still improve the performance of WCNN. Note that there are 2,148,000 NIR-VIS pairs in the testing. Hence a small improvement will result in the correct classification of many NIR-VIS pairs. These results highlight the importance of the Wasserstein distance and the low-rank constraint for the problems of sensing gap as well as over-fitting. When these two problems are well treated, deep learning methods could significantly improve NIR-VIS recognition performance.

\begin{table}[!t]
\renewcommand\arraystretch{1.3}
\caption{Rank-1 accuracy and verification rate on the Oulu-CASIA NIR-VIS Database. \label{tab:oulu}}
\centering
\begin{tabular}{|l|c|c|c|}
\hline
Methods & Rank-1 & FAR=1\%& FAR=0.1\%  \\ \hline\hline
MPL3 \cite{JChen:2009} & 48.9 & 41.9 & 11.4 \\ \hline
KCSR \cite{ZLei:2009}  & 66.0 & 49.7 & 26.1 \\ \hline
KPS \cite{Klare:2013}  & 62.2 & 48.3 & 22.2 \\ \hline
KDSR \cite{XHuang:2013}& 66.9 & 56.1 & 31.9 \\ \hline
H2(LBP3) \cite{MShao:2016} & 70.8 & 62.0 & 33.6 \\ \hline\hline
TRIVET\cite{XXLiu:2016} &92.2&67.9& 33.6\\ \hline
IDR  & 94.3& 73.4 & 46.2 \\\hline
IDR+low-rank & 95.0& 73.6 & 50.3 \\\hline
WCNN & 96.4 & 75.0 & 50.9 	\\\hline
WCNN + low-rank & \textbf{98.0} & \textbf{81.5} & \textbf{54.6}  \\\hline
\end{tabular}
\end{table}

\subsection{Results on the Oulu-CASIA NIR-VIS Database}
In this subsection, we evaluate the proposed methods on the Oulu-CASIA NIR-VIS Database. Compared to CASIA NIR-VIS 2.0 Database, the training set of the Oulu-CASIA NIR-VIS Database only consists of 20 subjects, which is of relative small-scale. Hence, it is challenging for a deep learning method due to over-fitting. We follow the testing protocol in \cite{MShao:2016} and compare WCNN with MPL3 \cite{JChen:2009}, KCSR \cite{ZLei:2009}, KPS \cite{Klare:2013}, KDSR \cite{XHuang:2013}, KDSR \cite{XHuang:2013}, H2(LBP3) \cite{MShao:2016} and TRIVET\cite{XXLiu:2016}. The results of MPL3, KCSR, KPS, KDSR, KDSR and H2(LBP3) are from \cite{MShao:2016}. TRIVET is used as the baseline of deep learning methods.

Table~\ref{tab:oulu} shows rank-1 accuracy and verification rates of different NIR-VIS matching methods. We observe that the methods can be nearly ordered in ascending rank-1 accuracy as MPL3, KPS, KCSR, KDSR, H2(LBP3), TRIVET, IDR, WCNN and WCNN+low-rank. The four deep learning methods perform significantly better than the five traditional methods in terms of rank-1 accuracy. Although the rank-1 accuracy of TRIVET is higher than that of H2(LBP3), VR@FAR=0.1\% of TRIVET is close to that of H2(LBP3). This may be because all VIS images of one subject are from the gallery and all their corresponding NIR images are treated as probe. Since NIR image and VIS image are paired during testing, it is easy for a deep learning method to give a high similarity score for paired data so that the rank-1 accuracy of one deep learning method is high. However, due to the sensing gap, a NIR image feature of one person is potentially similar to the VIS image feature of another person under the same expression. These two features may also have a higher similarity score so that verification rates of all methods are not very high at a low FAR. Due to the small-scale training set of this database, the four deep learning methods can not capture all variations so that their verification rates are lower than those on the CASIA NIR-VIS 2.0 Database. As expected, WCNN methods achieve the highest performance in terms of rank-1 accuracy and verification rates.

Fig.~\ref{fig:roc} (b) further plots the ROC curves of the four deep learning methods. The verification rates of all four methods drop dramatically as FAR becomes small. TRIVET obtains the lowest ROC curve. It is interesting to observe that there is only small improvement between the curves of WCNN and IDR. When the low-rank constraint is imposed on IDR, the ROC curve of IDR+low-rank is close to that of WCNN. This means that Wasserstein distance does not contribute too much to ROC curve. This is mainly because the training set of this database is small-scale so that WCNN over-fits on this small-scale training set. When low-rank constraint is imposed on the fully connected layer of WCNN, there is a significant difference between the ROC curves of WCNN and WCNN+low-rank. These results suggest that a suitable constraint on the fully connected layer can alleviate the over-fitting problem on a small training set.

\begin{table}[!t]
\renewcommand\arraystretch{1.3}
\caption{Rank-1 accuracy and verification rate on the BUAA NIR-VIS Database. \label{tab:buaa}}
\centering
\begin{tabular}{|l|c|c|c|}
\hline
Methods & Rank-1 & FAR=1\%& FAR=0.1\%  \\ \hline\hline
MPL3 \cite{JChen:2009} & 53.2 & 58.1 & 33.3 \\ \hline
KCSR \cite{ZLei:2009}& 81.4 & 83.8 & 66.7 \\ \hline
KPS \cite{Klare:2013} &66.6 & 60.2 & 41.7 \\ \hline
KDSR \cite{XHuang:2013} & 83.0 & 86.8 & 69.5 \\ \hline
H2(LBP3) \cite{MShao:2016} & 88.8&	88.8 & 73.4\\ \hline\hline
TRIVET\cite{XXLiu:2016} &93.9 & 93.0 & 80.9 \\ \hline
IDR \cite{RHe:2017} & 94.3 &93.4 & 84.7 \\\hline
IDR + low-rank & 94.8 & 94.5 & 86.0 \\\hline
WCNN & 95.4 & 93.9 & 86.9 	\\\hline
WCNN + low-rank & \textbf{97.4} & \textbf{96.0} & \textbf{91.9}  \\\hline
\end{tabular}
\end{table}

\subsection{Results on the BUAA VisNir Database}
In this subsection, we evaluate the proposed methods on the BUAA VisNir Database. As shown in Fig.~\ref{fig:face} (c), VIS and NIR images are well aligned and have similar appearance because they are captured simultaneously. These well-aligned NIR and VIS images potentially facilitate deep learning methods to capture intrinsic identity variation and reduce sensing gap. We follow the testing protocol in \cite{MShao:2016} to evaluate different NIR-VIS matching methods. The results for the BUAA VisNir database are presented in Table~\ref{tab:buaa} and Fig.~\ref{fig:roc} (c). The results of MPL3, KCSR, KPS, KDSR, KDSR and H2(LBP3) are from \cite{MShao:2016}.

We observe that the five deep learning methods perform better than the five traditional methods. The methods can be nearly ordered in ascending rank-1 accuracy as MPL3, KPS, KCSR, KDSR, H2(LBP3), TRIVET, IDR, IDR+low-rank, WCNN and WCNN+low-rank. Our WCNN+low-rank method improves the best rank-1 accuracy from 88.8\% to 97.4\% and VR@FAR=0.1 from 73.4\% to 91.9\%. When low-rank constraint and Wasserstein distance are introduced to IDR, IDR's performance is significantly improved. Particularly, the highest performance is achieved when both low-rank constraint and Wasserstein distance are used. This is because deep learning methods are simultaneously degraded by the sensing gap and the over-fitting problems. Our proposed architecture can naturally deal with these two problems in an end-to-end network, resulting in higher performance on this database.

\begin{figure}[t]
\center
    \subfigure[Without low-rank constraint]{\includegraphics[width=40mm]{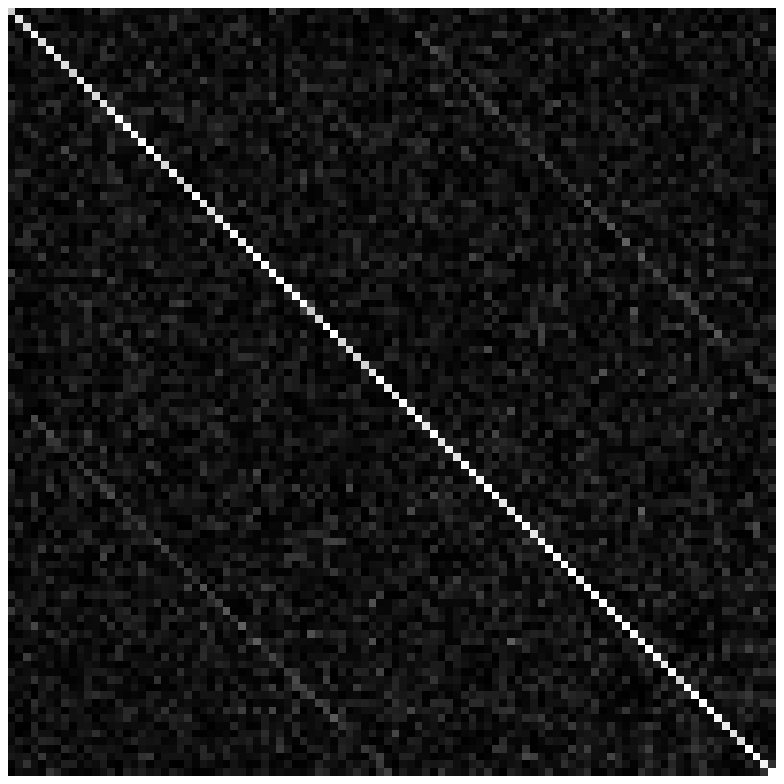}}
    \subfigure[With low-rank constraint]{\includegraphics[width=40mm]{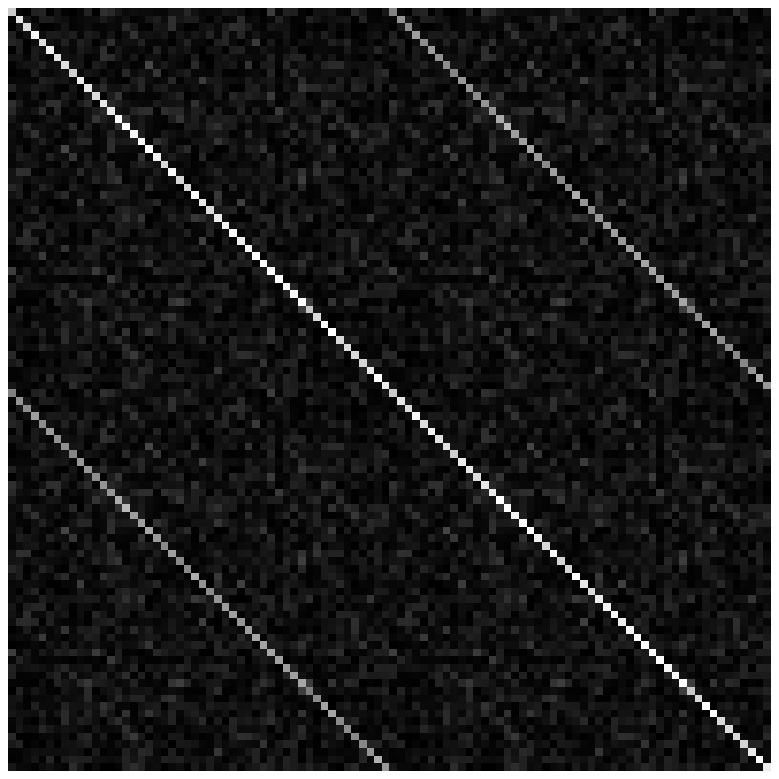}}
    \caption{A correlation illustration of the matrix $\mathrm{M}^T\mathrm{M}$ in the fully connected layer of WCNN. A lighter color indicates a higher correlation. When the low-rank correlation constraint is introduced, there is obvious variations on top-right and bottom-left areas of $\mathrm{M}^T\mathrm{M}$. \label{fig:corr}}
\end{figure}

From Fig.~\ref{fig:roc} (c), we observe that the methods can be nearly ordered in ascending ROC curve as TRIVET, IDR, IDR+low-rank, WCNN and WCNN+low-rank. The low-rank constraint significantly improves the ROC curves of IDR and WCNN especially when FAR is low. Since the training set of this database is of small-scale, deep learning may potentially over-fit on the training set. Fig.~\ref{fig:corr} further plots the values of the matrix $\mathrm{M}^T\mathrm{M}$ without (Fig.~\ref{fig:corr} (a)) or with (Fig.~\ref{fig:corr} (b)) the low-rank constraint on the fully connected layer of WCNN. A lighter color indicates a higher correlation. When the low-rank correlation constraint is used, there is obvious variations on top-right and bottom-left areas of $\mathrm{M}^T\mathrm{M}$. Note that $\mathrm{M}$ is composed of $F_N$ and $F_V$. The diagonal elements in the top-right and bottom-left areas have lighter color. This indicates that $F_N$ and $F_V$ are correlated, which reduces parameter space of the fully connected layer. These results further validate the effectiveness of the low-rank correlation constraint, suggesting the usage of correlation constraints on the fully connected layer to alleviate the over-fitting problem.





\section{Conclusion\label{sec:con}}
By naturally combining subspace learning and invariant feature extraction into CNNs, this paper has developed a Wasserstein CNN approach that uses only one network to map both NIR and VIS images to a compact Euclidean space. The high-level layer of WCNN is divided into two orthogonal subspaces that contain modality-invariant identity information and modality-variant light spectrum information, respectively. Wasserstein distance has been used to measure the difference between heterogeneous feature distributions and proven to be effective to reduce the sensing gap. To the best of our knowledge, it is the first attempt in NIR-VIS field to formulate a probability distribution learning for VIS-NIR matching. In addition, low-rank constraint has been studied to alleviate the over-fitting problem on small-scale NIR-VIS face data. An alternating minimization approach has been developed to minimize the joint formulation of WCNN in an end-to-end way. Experimental results on three challenging NIR-VIS face recognition databases show that our WCNN methods significantly outperform state-of-the-art NIR-VIS face recognition methods.



%

%
%
%
%
%

\ifCLASSOPTIONcaptionsoff
  \newpage
\fi



{
\bibliographystyle{IEEEtran}
\bibliography{reference}
}
\end{document}